\documentclass[nolinenumbers]{iucrjournals}
\usepackage{amsmath,amssymb,amsfonts}
\usepackage{algorithmic}
\usepackage{algorithm}
\usepackage{array}
\usepackage{bm}
\usepackage{textcomp}
\usepackage{stfloats}
\usepackage{url}
\usepackage{verbatim}
\usepackage{graphicx}
\usepackage{textcomp}
\usepackage{xcolor}
\usepackage{dsfont}
\usepackage{bm,mathabx,amsthm}
\usepackage{subcaption}
\usepackage{graphicx}
\usepackage{booktabs}
\usepackage{romannum}

\title{OM\textsuperscript{2}P: Offline Multi-Agent Mean-Flow Policy}

\author[1]{Zhuoran Li}%
\author[1]{Xun Wang}%
\author[1]{Hai Zhong}%
\author[2]{Qingxin Xia}
\author[2]{Lihua Zhang}
\author[1]{Longbo Huang\thanks{{Corresponding Author}}}

\affil[1]{Institute for Interdisciplinary Information Sciences (IIIS), Tsinghua University \{lizr20,wang-x24,zhongh22\}@mails.tsinghua.edu.cn, longbohuang@tsinghua.edu.cn }
\affil[2]{ByteDance Inc. \{xiaqingxin,lizhiyu.0\}@bytedance.com}

\begin{document} 

\pagenumbering{arabic}
\maketitle

\begin{abstract}
Generative models, especially diffusion and flow-based models, have been promising in offline multi-agent reinforcement learning. However, integrating powerful generative models into this framework poses unique challenges. In particular, diffusion and flow-based policies suffer from low sampling efficiency due to their iterative generation processes, making them impractical in time-sensitive or resource-constrained settings. To tackle these difficulties, we propose Offline Multi-Agent Mean-Flow Policy (\textbf{OM\textsuperscript{2}P}), a novel offline MARL algorithm to achieve efficient \emph{one-step} action generation. To address the misalignment between generative objectives and reward maximization, we introduce a reward-aware optimization scheme that integrates a carefully-designed mean-flow matching loss with Q-function supervision. Additionally, we design a generalized timestep distribution and a derivative-free estimation strategy to reduce memory overhead and improve training stability. Empirical evaluations on Multi-Agent Particle and MuJoCo benchmarks demonstrate that OM\textsuperscript{2}P achieves superior performance, with up to a $3.8\times$ reduction in GPU memory usage and up to a $10.1\times$ speed-up in training time. Our approach represents the first to successfully integrate mean-flow model into offline MARL, paving the way for practical and scalable generative policies in cooperative multi-agent settings.
\end{abstract}


\section{Introduction}
Offline Multi-Agent Reinforcement Learning (Offline MARL) focuses on acquiring coordinated policies from fixed datasets without further environment interaction. This setup is particularly valuable in risk-sensitive or limited-interaction domains such as autonomous driving \cite{zhang2024multiagentreinforcementlearningautonomous}, robotic manipulation \cite{feng2025learningmultiagentlocomanipulationlonghorizon}, and distributed resource allocation \cite{hady2025multiagentreinforcementlearningresources}, where data collection is expensive or unsafe. Recently, generative models—especially diffusion and flow-based models—have shown promise for offline policy learning by modeling expressive, multimodal action distributions. These models, originally developed for image generation \cite{ho2020denoising,songscore,lipmanflow}, have been extended to offline RL \cite{wangdiffusion,chenoffline,park2025flowqlearning} and further adapted to multi-agent settings for action generation \cite{li2023conservatismdiffusionpoliciesoffline}, trajectory planning \cite{zhu2024madiff}, and data augmentation \cite{fu2025ins}.

However, their reliance on iterative sampling often leads to inefficiencies in training and inference, posing a bottleneck for large-scale or time-sensitive applications. 
While these generative models offer strong representational capacity, their sampling inefficiency poses a major obstacle in offline MARL. Unlike conventional policy networks, these policies require multiple iterative steps to sample actions, leading to high computational overhead for training and inference. 
This issue is magnified in multi-agent scenarios due to the need for repeated joint action sampling across agents, which significantly amplifies the computational burden. Improving the sampling efficiency of diffusion or flow policies is therefore critical for their practical deployment in time-sensitive, real-world multi-agent applications \cite{xie2025multiuavformationcontrolstatic}. 

Recent advances in flow-based models, particularly mean-flow model \cite{geng2025mean}, have demonstrated the potential for efficient one-step sampling by replacing instantaneous velocity fields with their averaged counterparts, enabling fast and high-quality generation in image synthesis.
Inspired by mean-flow model, 
we thus explore whether they can serve as policy networks in offline MARL to resolve the efficiency bottleneck by avoiding multi-step inference during training, especially policy distillation \cite{park2025flowqlearning}. However, directly applying this adaptation poses key challenges: (i) The training objectives of mean-flow model and offline MARL are inherently misaligned: maximizing expected rewards is not equivalent to minimizing negative log-likelihood. (ii) Computing the gradients of the mean-flow velocity, which is required by the partial derivation of the mean-flow objective, introduces significant computational overhead. (iii) Existing mean-flow training formulations are not tailored for one-step policy learning, especially with uniform timestep sampling, which is critical in RL applications. 

Motivated by the limitations of generative models in offline MARL and the burgeoning development of flow-based models, we ask: \textbf{\textit{Is it possible to design ultra-efficient training and inference flow-based policies in offline MARL?}}

To address the efficiency challenges in offline MARL, we introduce \textit{Offline Multi-Agent Mean-Flow Policy} (OM\textsuperscript{2}P), a novel framework that seamlessly incorporates mean-flow model as policy networks for offline MARL. OM\textsuperscript{2}P redefines the training paradigm of mean-flow-based policy networks via generalized timestep sampling and derivative-free target mean-velocity estimation to enable efficient one-step action sampling without relying on distillation and iterative action generation.
Our key contributions are summarized as follows: 

\begin{itemize}
    \item We introduce a novel framework of the mean-flow-based model into offline MARL, avoiding policy distillation to achieve one-step action generation. This design significantly improves training and inference efficiency while maintaining strong policy expressiveness. 
    
    \item We propose a decentralized offline training scheme that leverages a modified mean-flow matching loss in conjunction with Q-function supervision. To further improve efficiency, we adopt a generalized exponential-family distribution for timestep sampling and replace the exact gradient of target mean-velocity with a numerically stable finite-difference approximation, leading to reduced memory overhead and faster optimization.
    
    \item We conduct comprehensive experiments on standard offline MARL benchmarks, including Multi-Agent Particle (MPE) and Multi-Agent MuJoCo (MAMuJoCo) environments. OM\textsuperscript{2}P consistently achieves near-optimal performance with up to $3.8\times$ GPU memory reduction and $10.1\times$ speed up in training time, demonstrating both its efficiency and effectiveness.
\end{itemize}

\section{Related Work}

\subsection{Offline MARL}
Offline Multi-Agent Reinforcement Learning (Offline MARL) faces unique challenges such as distributional shift, credit assignment, and multi-agent coordination under fixed datasets \cite{levine2020offlinereinforcementlearningtutorial,formanek2023off}. 
Existing works extend pessimistic value estimation from single-agent RL to the multi-agent setting (e.g., MAICQ \cite{yang2021believe}, MABCQ \cite{jiang2023offline}, CFCQL \cite{shao2023counterfactual}, and OMAC \cite{wang2023offline}), primarily focusing on conservative learning principles.
Other methods, for instance, OMAR \cite{pan2022plan}, SIT \cite{Tian_Kuang_Liu_Wang_2023} and InSPO \cite{Liu_Lin_Yu_Wu_Liang_Li_Ding_2025} address local optima and data imbalance through actor rectification, reliable credit assignment and sequentially optimizing agent policies in-sample to avoid out-of-distribution joint actions as a potential risk. 

Recent methods leverage diffusion models to solve data scarcity and multi-modality in offline MARL, achieving coordinated behavior through trajectory modeling (MADiff \cite{zhu2024madiff}, DoF \cite{lidof2025}), interaction-aware data synthesis (INS \cite{fu2025ins}), diffusion-based action generation (DOM2 \cite{li2023conservatismdiffusionpoliciesoffline}), and decentralized policy learning with score function decomposition (OMSD \cite{qiao2025offlinemultiagentreinforcementlearning}).
Planning-based and world-model approaches (e.g., MOMA-PPO \cite{barde2024momappo}, AlberDICE \cite{Matsunaga2023AlberDICE}) further improve coordination under offline constraints, while trajectory generation methods (e.g.,  MAT \cite{wen2022multi}, MADT \cite{meng2023offline}, and MADTKD \cite{tseng2022offline}) enhance policy learning by modeling the temporal dependencies and structure of agent behaviors.

\subsection{Diffusion Models and Flow Matching in Offline RL}

Diffusion-based methods have been widely adopted in offline RL for modeling action distributions as policies (Diff-QL \cite{wangdiffusion}, SfBC \cite{chenoffline}, CEP \cite{lu2023contrastive}, EDP \cite{kang2023efficient}, and IDQL \cite{hansen2023idql}), behavior cloning (Diff-BC \cite{pearceimitating}, ARQ \cite{goo2022knowboundariesnecessityexplicit}), trajectory planning (Diffuser \cite{janner2022planning}, Decision Diffuser \cite{ajayconditional}), and data augmentation (SynthER \cite{lu2023synthetic}, GTA \cite{lee2024gta}). 
The emergence of diffusion policies has also inspired extensions to robotic learning, e.g., Diffusion Policy \cite{chi2023diffusionpolicy} and DPPO \cite{ren2024diffusion}.
In parallel, flow-based approaches, e.g., FQL \cite{park2025flowqlearning}, FlowQ \cite{alles2025flowqenergyguidedflowpolicies}, QIPO \cite{zhangenergy}, and SSCP \cite{koirala2025flowbasedsinglestepcompletionefficient} offer efficient alternatives to iterative diffusion by enabling action sampling via flow matching or energy-weighted training.

Different from prior diffusion-based and flow-based approaches that rely on multi-step denoising or distillation—which can be computationally prohibitive in multi-agent settings—OM\textsuperscript{2}P introduces a mean-flow policy that enables efficient one-step action generation. By seamlessly integrating a generalized mean-flow model into offline MARL, our method eliminates the need for multi-step action sampling while naturally aligning policy learning with value estimates. In particular, the mean-flow generator is supervised by the target mean velocity derived from the offline Q-function, providing a stable and principled form of value alignment. This unified design not only maintains high efficiency and low memory overhead, but also proves sufficiently expressive across benchmarks, obviating the need for additional energy-based guidance or more complex reward alignment strategies and offering a scalable and efficient solution for high-dimensional multi-agent coordination.

\section{Background}

\subsection{Offline MARL}

We consider fully cooperative multi-agent tasks modeled as a decentralized partially observable Markov decision process (Dec-POMDP) \cite{oliehoek2016concise}, defined by a tuple $G = \langle \mathcal{I}, \mathcal{S}, \mathcal{O}, \mathcal{A}, \Pi, \mathcal{P}, \mathcal{R}, n, \gamma \rangle$. Here, $\mathcal{I} = \{1, \ldots, n\}$ denotes the set of $n$ agents, $\mathcal{S}$ is the global state space, and $\mathcal{O} = (\mathcal{O}_1, \ldots, \mathcal{O}_n)$ represents the local observation spaces for each agent. Each agent $i$ selects actions $a_i \in \mathcal{A}_i$, where $\mathcal{A} = (\mathcal{A}_1, \ldots, \mathcal{A}_n)$ denotes the joint action space. 
The environment evolves according to the dynamics function \({P}(s'|s, a): \mathcal{S} \times \mathcal{A} \to \mathcal{S}\), where \({P}(s'|s, a)\in\mathcal{P}\) denotes the probability of transitioning to state \(s'\) given the current state \(s\) and action \(a\). Each agent receives an observation \(o_i \in \mathcal{O}_i\) and a scalar reward \(r_i \in \mathbb{R}\) from a shared reward function \(\mathcal{R}: \mathcal{S} \times \mathcal{A} \to \mathbb{R}\).
Within the framework of cooperative MARL, the agents' collective endeavor is to devise a set of policies \(\bm{\pi} = \{\pi_1, \ldots, \pi_n\}\in\Pi\) to maximize the expected cumulative discounted rewards:
\(
\mathbb{E}_{\bm{\pi}} \left[ \sum_{t=0}^{\infty} \gamma^t \sum_{i=1}^n r_i^t \right]
\),
where \(\gamma \in [0,1)\) represents the discount factor.
In the offline MARL paradigm, agents are trained exclusively on a pre-collected dataset \(\mathcal{D}\) generated by sampling from a potentially unknown behavior policy \(\boldsymbol{\pi}_{\beta}\), thereby decoupling the training process from direct interaction with the environment.

\subsection{Flow Matching}

To pave the way for introducing our OM\textsuperscript{2}P algorithm, we provide a brief overview of flow matching and mean-flow model, which serve as foundational tools in our design.

Flow matching~\cite{lipmanflow,liuflow,albergobuilding} is a framework of generative model that achieves efficient and high-quality sample generation by learning flow trajectories from a simple base distribution (e.g., Gaussian) to a complex target distribution~\cite{esser2024scaling,lipman2024flow}.
Given data \( x \sim p(x) \) over \( \mathbb{R}^d \), it constructs a trajectory \( x_t=\psi(t, x) \) that transforms noise samples \( \psi(0, x) \sim \mathcal{N}(0, I_d) \) into data samples \( \psi(1, x) = x \sim p(x) \), where the velocity is defined as \( v(t, x) = \frac{\mathrm{d}}{\mathrm{d}t} \psi(t, x) \). 
To learn this transformation, the model optimizes the velocity \( v_{\theta}(t, x) \) using the objective:
$\min_{\theta} \mathbb{E}_{x_0 \sim \mathcal{N}(0, I_d),\, x_1 \sim p(x),\, t \sim U([0,1])} \left\| v_{\theta}(t, x_t) - v(t,x_t) \right\|_2^2.$
After training, new samples can be generated by numerically integrating the learned velocity field via the ODE: $x_t = x_r + \int_r^t v_{\theta}(\tau, x_{\tau})\, \mathrm{d}\tau.$ 

The mean-flow framework~\cite{geng2025mean} simplifies the sampling by replacing instantaneous velocities with a mean velocity: $u(x_r, r, t)=\frac{1}{t - r} \int_{r}^{t} v(\tau, x_\tau) \, \mathrm{d}\tau,$
yielding the closed-form update \( x_t = x_r + u(x_r, r, t)(t - r) \). This formulation avoids expensive numerical integration, enabling efficient single- or multi-step sampling. To model the mean velocity, we adopt a neural network \( u_{\theta}(x_r, r, t) \) trained to satisfy the mean-flow identity (equals to Equation (9) in \cite{geng2025mean}) via:
\begin{equation}
\label{eq:mfloss}
\mathcal{L}(\theta) = \mathbb{E} \left[ \left\| u_{\theta}(x_r, r, t) - \text{stopgrad}(u_{\text{target}}) \right\|_2^2 \right],
\end{equation}
where \( x_r = (1 - r)x_0 + r x_1 \), with \( x_0 \sim \mathcal{N}(0, I_d) \), \( x_1 \sim p(x) \), and \( r, t \sim U([0, 1]) \). The target velocity is computed as \( u_{\text{target}} = v(r, x_r) - (r - t)\, \frac{\mathrm{d}}{\mathrm{d}r} u(x_r, r, t) \), where the temporal derivative is approximated by \( \frac{\mathrm{d}}{\mathrm{d}r} u = v(r, x_r)\, \partial_{x_r} u + \partial_r u \).

\subsection{Bridge to Mean-Flow Policy}

Flow matching trains a velocity field along interpolation paths between noisy inputs and dataset actions, enabling sample generation via denoising. While effective for static data modeling, directly applying this paradigm to offline MARL is inadequate due to key mismatches in objective, efficiency, and scalability.

Unlike generative models, offline MARL demands policies that maximize cumulative rewards, not just replicate dataset distributions—rendering vanilla flow objectives misaligned \cite{wangdiffusion}. Moreover, multi-step generation undermines training efficiency, especially in multi-agent systems where joint action sampling is costly \cite{zhu2024madiff}. Although mean-flow model supports one-step generation, their objectives are not inherently designed for it, and computing target velocity fields remains expensive and reward-agnostic \cite{hang2024improvednoiseschedulediffusion}. These issues are magnified in decentralized, multi-agent scenarios \cite{xie2025multiuavformationcontrolstatic}. 

To address these challenges, we propose OM\textsuperscript{2}P—a novel offline MARL framework that repurposes mean-flow model as an efficient, reward-aligned policy network. Through a carefully designed objective and optimization strategy, OM\textsuperscript{2}P enables fast one-step action generation while preserving policy expressiveness and improving training efficiency.

\begin{figure*}[ht]
    \centering
    \includegraphics[width=1.0\linewidth]{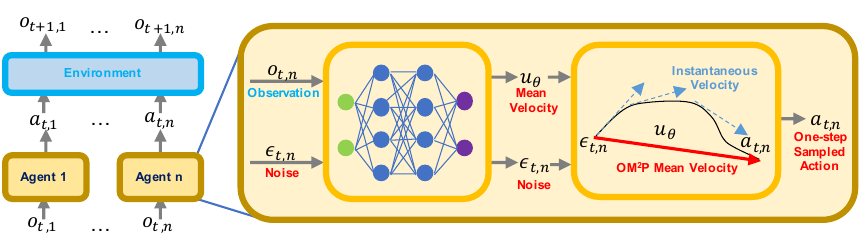}
    \vspace{-0.4in}
    \caption{Overview of the decentralized OM\textsuperscript{2}P framework. OM\textsuperscript{2}P illustrates a single representative agent performing scalable, one-step action generation to avoid the computational bottleneck of iterative sampling inherent to multi-agent settings.}
    \label{fig:om2p}
\end{figure*}

\section{The OM²P Algorithm }\label{chap:algorithms}

We present OM\textsuperscript{2}P, a novel algorithm addressing the efficiency and scalability bottlenecks of generative policies in multi-agent settings.
By leveraging a decentralized mean-flow architecture (as illustrated by the representative agent in Figure~\ref{fig:om2p}), OM\textsuperscript{2}P enables fast one-step action generation, avoiding the computationally prohibitive iterative sampling and distillation typical of single-agent adaptations.
This framework introduces specific modeling and optimization innovations to meet the distinct demands of offline MARL.
In the following sections, we begin by detailing the integration of mean-flow model into offline MARL and highlighting the inefficiencies of naively applying standard mean-flow formulations.
We then introduce a novel training objective tailored to efficient reward-aware policy learning. Finally, we provide an overview of the full OM\textsuperscript{2}P algorithm and discuss its advantages.

\subsection{Mean-Flow Model in Offline MARL}

We begin by describing how to seamlessly integrate mean-flow model into the offline multi-agent reinforcement learning (MARL) setting and highlight the inefficiencies that arise from directly applying standard formulations. Specifically, we instantiate the mean-flow model as the policy network for each agent in a decentralized MARL framework (for clarity, we omit the agent index in subsequent notation). That is, the conditional policy \(\pi_{\theta}(a|o)\), which maps local observations to action distributions, is parameterized via a mean-flow model.

Given the observation-action pairs \((o, a) \sim \mathcal{D}\), which are sampled from a fixed dataset induced by a behavior policy \(\pi_\beta\), the mean-flow policy constructs an action trajectory \(a_t = \psi(t, a|o)\) using the instantaneous velocity \(v(a_r, r|o) = \frac{\text{d}a_r}{\text{d}r}\) and the mean velocity \(u(a_r, r, t|o) = \frac{1}{t - r} \int_r^t v(a_\tau, \tau|o) \, \text{d}\tau\). This defines a continuous path from the noise \(a_0 = \psi(0, a|o) \sim \mathcal{N}(0, I_d)\) to data \(a_1 = \psi(1, a|o) \sim \pi_\beta(a|o)\), where the intermediate states satisfy \(a_t = a_r + \int_r^t v(a_\tau, \tau|o) \, \text{d}\tau = a_r + (t - r) u(a_r, r, t|o)\).

We use the neural network $u_{\theta}(a_r,r,t|o)$ to represent the policy $\pi_{\theta}(a|o)$ for action generation. A natural extension of mean-flow model~\cite{geng2025mean} is to train a velocity field $u_{\theta}(a_r, r, t|o)$ that maps interpolated the noisy actions to their expected displacement: 
\begin{equation}
    \label{eq:bcloss}
    \mathcal{L}_{bc}(\theta)=\mathbb{E}[\Vert u_\theta(a_r,r,t|o) - \text{stopgrad}(u_{\text{target}})\Vert^2].
\end{equation}

We sample $(o,a)\sim\mathcal{D}$ from the fixed dataset, $(t,r)\sim U([0,1])$ uniformly, and set $a_r = (1{-}r)\epsilon + ra$ with $\epsilon \sim \mathcal{N}(0,I)$ to interpolate between the action and Gaussian noise for simplicity as a common practice \cite{geng2025mean,lipman2024flow}. The regression target is $u_{\text{target}} = v(a_r,r|o) - (r{-}t)\frac{\text{d}}{\text{d}r}u_\theta(a_r,r,t|o)$, where the temporal derivative is given by $\frac{\text{d}}{\text{d}r}u_\theta = v(a_r,r|o)\partial_{a_r}u_\theta + \partial_r u_\theta$. We apply $\text{stopgrad}(\cdot)$ to prevent double backpropagation and stabilize training. Based on Equation~\eqref{eq:bcloss}, we train the mean velocity field to approximate the behavior policy $\pi_{\beta}$. At inference, actions are generated via the update rule:
\begin{equation}
\label{eq:sample}
a_t = a_r + (t - r) u_\theta(a_r, r, t|o),
\end{equation}
starting from $a_0 \sim \mathcal{N}(0, I)$ until $t=1$. For one-step sampling, we compute $\tilde{a} = a_0 + u_\theta(a_0, 0, 1|o)$ using a single network evaluation.

However, directly applying this mechanism to offline MARL faces two limitations: (i) Training Instability and Inefficiency: Standard mean-flow losses require estimating target velocities through both forward and backward passes, involving gradients w.r.t.\ $a_r$ and $t$. This leads to unstable supervision, inaccurate gradients, and high memory consumption in multi-agent settings. (ii) {Temporal Mismatch:  Uniform timestep sampling treats all $t \in [0, 1]$ equally, while our goal is achieving accurate one-step action generation (i.e., $t = 1$). This mismatch degrades the performance of generating actions. 
Furthermore, the objective mismatch—where offline MARL aims to maximize the expected episodic return, while mean-flow model focuses on fitting the data distribution—can lead to suboptimal policy learning by failing to emphasize reward-relevant behaviors.

To overcome these challenges, OM\textsuperscript{2}P innovatively introduces \emph{adaptive timestep distribution} and \emph{derivative-free velocity estimation} for more efficient mean-flow generation. Additionally, it leverages Q-value guidance to surpass the behavior policy \cite{wangdiffusion,park2025flowqlearning}. Attributed to these elaborate designs, our proposed OM\textsuperscript{2}P consistently outperforms existing baselines.

\subsection{Key Components of OM²P }
We now outline the essential components of OM\textsuperscript{2}P: an adaptive objective enabled by generalized timestep sampling, derivative-free velocity estimation, and Q-value guidance for policy improvement.

\paragraph{Generalized Timestep Sampling.}

Instead of \cite{lipman2024flow,geng2025mean}, 
we introduce a generalized timestep distribution:
\begin{equation}
\label{eq:gendis}
p(t; \xi) = \frac{1}{Z(\xi)} \exp(\xi^\top h(t)),
\end{equation}

\noindent
where $\xi$ is a fixed coefficient vector that parameterizes and thereby determines the functional form of the timestep distribution, $Z(\xi)$ is the normalizing constant, and $h(t) = [\log t, \log(1 - t), t, 1]^\top$ is a polynomial feature basis. 
Moreover, we set \( r = 1 - t \) to ensure that the distribution of \( t \) also governs that of \( r \), enabling coupled control over the interpolation and learning emphasis.

This formulation generalizes hand-crafted schedules (e.g., Beta, exponential) via a reparameterized timestep distribution $p(t;\xi)$ compared with a default uniform distribution in \cite{lipman2024flow,geng2025mean}, enabling task-adaptive one-step generation by controlling $\xi$ values, with $h(t)$ including polynomial and logarithmic terms for flexible shaping without explicitly computing the normalizing constant $Z(\xi)$.
Focusing more on informative timesteps (e.g., near $t{=}1$) improves gradient quality and accelerates policy learning.
As shown in Figure~\ref{fig:generalized_bc} (left), using a generalized distribution ($\xi=[5,5,0,0]$) improves stability over the uniform case ($\xi=[0,0,0,0]$) under pure behavior cloning, demonstrating that more flexible timestep distributions better emphasize informative regions and enhance learning.

\paragraph{Derivative-Free Velocity Estimation.}

To mitigate the high memory cost and instability of gradient-based calculation, we introduce a derivative-free velocity estimation approach that circumvents partial derivative computation while maintaining training effectiveness. We avoid costly backpropagation through intermediate $r$ by estimating the time derivative via finite differences:
\begin{equation}
\label{eq:difference}
\frac{\text{d} u_\theta}{\text{d}r} \approx \frac{u_\theta(a_{r+\Delta r}, r+\Delta r, t|o) - u_\theta(a_r, r, t|o)}{\Delta r}.
\end{equation}
Here the calculation of $a_{r+\Delta r}$ and $a_{r}$ follows the same interpolation path for consistency. By avoiding partial derivatives of the mean velocity and using a forward-only finite difference approximation, our derivative-free method eliminates second-order gradient tracking, substantially reduces memory consumption, and enhances stability and computational efficiency in high-dimensional settings.
As illustrated in Figure~\ref{fig:generalized_bc} (right), when $\Delta r \leq 10^{-8}$ as a float32 value, the finite-difference estimator yields stable performance nearly indistinguishable from that using exact gradients. This empirical observation validates the reliability of our derivative-free approach within practical numerical ranges.

\begin{figure}[htp]
    \centering
    \includegraphics[width=\linewidth]{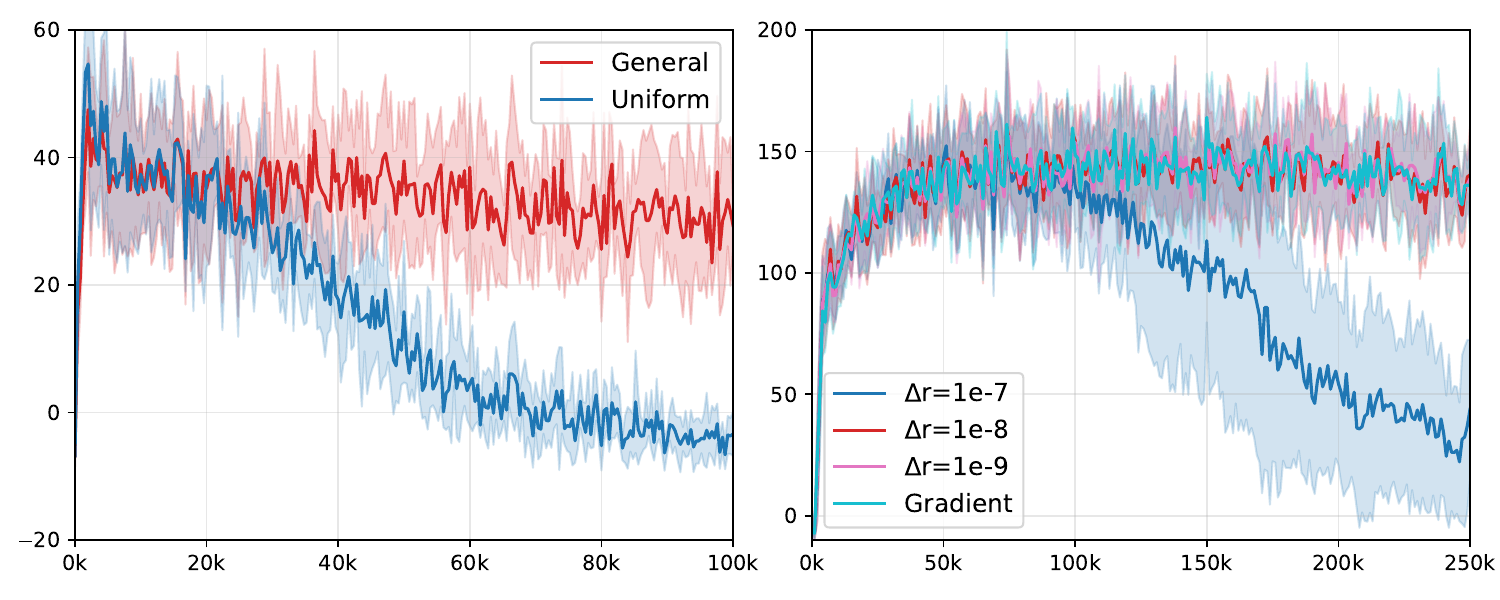}
    \vspace{-0.3in}
    \caption{ Average episodic return measured at different training timesteps in MPE World task via expert dataset.
\textbf{Left:} Behavior cloning performance using a Beta distribution ($\xi = [5,5,0,0]$) vs. uniform ($\xi = [0,0,0,0]$). Non-uniform weighting improves stability by emphasizing critical timesteps. 
\textbf{Right:} Performance under varying finite-difference step sizes $\Delta r$ for estimating $\frac{\text{d}u_{\theta}}{\text{d}r}$. Values $\Delta r \leq 10^{-8}$ yield results comparable to exact gradients, validating the reliability of our approximation. Details are shown in the Appendix~\ref{appendix:expresults}.
}
    \label{fig:generalized_bc}
\end{figure}

\vspace{-0.2in}

\paragraph{Policy Improvement of One-Step Mean-Flow Policy.}

While our mean-flow framework effectively models action generation, optimizing only the behavior cloning loss via Equations~\eqref{eq:bcloss}, \eqref{eq:gendis}, and \eqref{eq:difference} is insufficient for offline MARL. These objectives merely encourage imitation of the dataset behavior, without driving improvement. Even with refined timestep modeling and derivative-free enhancements, the policy remains bound to suboptimal actions unless guidance by reward signals.

To address this limitation, we incorporate a value-based objective that complements the mean-flow loss with Q-value guidance, following the principle that actions should not only match the data distribution but also yield higher expected returns. Specifically, we define the policy loss as:
\begin{equation}
    \label{eq:policyloss}
    \begin{aligned}
        \mathcal{L}(\theta) = \mathcal{L}_{\text{bc}}(\theta) + \mathcal{L}_q(\theta) = \mathcal{L}_{\text{bc}}(\theta) - \eta \mathbb{E}[Q_{\phi}(o, \tilde{a})],
    \end{aligned}
\end{equation}
where $(o, a) \sim \mathcal{D}$ is drawn from the offline dataset, and $\tilde{a}$ is sampled from the mean-flow policy. The Q-guided term promotes actions with higher estimated returns, enabling the policy to depart from the behavior distribution and favor more rewarding choices.

To ensure effective guidance, the critic is trained following \cite{park2025flowqlearning}:
\begin{equation}
    \label{eq:criticloss}
    \mathcal{L}_{\phi} = \mathbb{E}\left[\left(r + \gamma \, \text{mean}_{j=1,2} Q_{\phi_j'}(o', \tilde{a}') - Q_{\phi}(o, a)\right)^2\right],
\end{equation}
where $(o, a, r, o') \sim \mathcal{D}$, $\tilde{a}'$ is sampled from the mean-flow policy via one-step action sampling and $\gamma$ is the discounted factor. The integration of reward-based supervision with smooth action modeling facilitates performance beyond behavior cloning.

\subsection{OM²P Algorithm}

\begin{algorithm}[ht]
   \caption{Offline Multi-Agent Mean-Flow Policy (OM\textsuperscript{2}P)} 
   \label{alg:om2p}
\begin{algorithmic}[1]
   \STATE {\bfseries Input:} Initialize Q-networks $Q_{{\phi}_j}^1,Q_{{\phi}_j}^2$, policy network $\pi_j$ with random parameters ${\phi}_j^1,{\phi}_j^2,{\theta}_j$, target networks with $\overline{{\phi}}_j^1\leftarrow{\phi}_j^1,\overline{{\phi}}_j^2\leftarrow{\phi}_j^2,\overline{{\theta}}_j\leftarrow{\theta}_j$ for each agent $j=1,\dots,N$ and dataset $\mathcal{D}$. \textbf{// $\text{Initialization}$}\label{algs:initialization}
   \FOR{training step $t=1$ {\bfseries to} $T$}
   \FOR{agent $j=1$ {\bfseries to} $n$}
   \STATE Sample a random minibatch of $\mathcal{S}$ samples $({o}_j,{a}_j,{r}_j,{o}'_j)$ from dataset $\mathcal{D}'$. \textbf{// $\text{Sampling}$}\label{algs:sampling}
   \STATE Update critics ${\phi}_j^1,{\phi}_j^2$ to minimize \eqref{eq:criticloss}. \textbf{// $\text{Update\  Critic}$}\label{algs:updatecritics}
   \STATE Update the actor ${\theta}_j$ to minimize \eqref{eq:policyloss}. \textbf{// $\text{Update\ Mean-Flow Policy}$} \label{algs:updateactor} 
   \STATE Update target networks: $\overline{{\phi}}_j^k\leftarrow\rho{\phi}_j^k+(1-\rho)\overline{{\phi}}_j^k$,$(k=1,2)$,$\overline{{\theta}}_j\leftarrow\rho{\theta}_j+(1-\rho)\overline{{\theta}}_j$. \label{algs:softupdate}
   \ENDFOR
   \ENDFOR
\end{algorithmic}
\end{algorithm}

Summarizing the above, we present the pseudo-code of OM\textsuperscript{2}P in Algorithm~\ref{alg:om2p}. It constitutes a decentralized offline MARL framework, where each agent $i=1,\dots,N$ maintains its own critic networks $Q_{\phi_1}^i, Q_{\phi_2}^i$ and a policy network $\pi_{\theta}^i$.
Line~\ref{algs:initialization} initializes all critic and actor networks along with their respective target counterparts. Line~\ref{algs:sampling} performs mini-batch sampling from the offline dataset $\mathcal{D}$ to construct training data. The critic networks are updated in Line~\ref{algs:updatecritics} by minimizing a Bellman regression loss (Equation~\eqref{eq:criticloss}), ensuring reliable value estimation. Subsequently, Line~\ref{algs:updateactor} updates the actor networks by minimizing the regularized mean-flow objective (Equation~\eqref{eq:policyloss}), integrating both behavioral fidelity and reward supervision. Line~\ref{algs:softupdate} applies soft update to the target networks, stabilizing the learning process.

OM\textsuperscript{2}P presents a novel integration of mean-flow generative modeling into decentralized offline MARL, offering an efficient solution for one-step action generation without policy distillation. To emphasize its practical advantages, we summarize the framework’s principal strengths below:

\textbf{Efficient policy learning.}
In contrast to iterative diffusion- and flow-based methods (e.g., Diff-QL \cite{wangdiffusion}, SfBC \cite{chenoffline}), OM\textsuperscript{2}P achieves efficient one-step synthesis via a unified objective that strictly couples adaptive mean-flow matching with reward-aware supervision. This tailored integration aligns generative modeling with value maximization, ensuring that policy learning remains grounded in the data distribution while significantly reducing sampling and computational overheads.

\textbf{Stable training with reduced computation.}
To overcome the convergence instability of standard flow models in RL, OM\textsuperscript{2}P employs generalized timestep sampling and a derivative-free velocity approximation to eliminate costly second-order gradient computations. Different from FQL \cite{park2025flowqlearning} that rely on complex rollouts and distillation, our native one-step design enhances training stability and efficiency, offering a specialized solution for the high computational demands of multi-agent environments.

\textbf{Modular and scalable architecture.}
OM\textsuperscript{2}P features a modular architecture where each component—optimized for stability or efficiency—avoids additive complexity. Computational costs scale linearly with agents, limited to target estimation and single-pass inference. By bypassing multi-step sampling and distillation pipelines, OM\textsuperscript{2}P ensures scalability, utilizing value guidance within the mean-flow framework to maintain robust performance and mitigate risks in high-dimensional offline MARL.

\begin{table*}[ht]
\begin{center}
\begin{tabular}{ccccc}
\toprule
Predator–Prey&OMAR&MA-SfBC&MA-FQL&OM\textsuperscript{2}P\\
\midrule
Medium-Replay&{86.8$\pm$43.7}&26.1$\pm$10.0&\textbf{203.8$\pm$43.3}&\textbf{203.1$\pm$59.3}\\
Medium&116.9$\pm$45.2&127.0$\pm$50.9&{255.1$\pm$65.0}&\textbf{257.3$\pm$52.0}\\
Medium-Expert&128.3$\pm$35.2&152.3$\pm$41.2&\textbf{233.0$\pm$42.5}&230.2$\pm$71.1\\
Expert&202.8$\pm$27.1&256.0$\pm$26.9&{289.5$\pm$28.6}&\textbf{330.3$\pm$40.9}\\
\midrule
Avg.&133.7$\pm$37.8&140.4$\pm$32.3&245.4$\pm$44.9&\textbf{255.2$\pm$55.9}\\
\midrule
World&OMAR&MA-SfBC&MA-FQL&OM\textsuperscript{2}P\\
\midrule
Medium-Replay&21.1$\pm$15.6&9.1$\pm$5.9&\textbf{107.0$\pm$30.0}&\textbf{106.0$\pm$10.2}\\
Medium&45.6$\pm$16.0&54.2$\pm$22.7&136.1$\pm$18.5&\textbf{142.0$\pm$18.6}\\
Medium-Expert&71.5$\pm$28.2&60.6$\pm$22.9&120.7$\pm$16.9&\textbf{126.9$\pm$30.2}\\
Expert&84.8$\pm$21.0&97.3$\pm$19.1&139.4$\pm$13.9&\textbf{175.0$\pm$21.5}\\
\midrule
Avg.&55.8$\pm$20.2&55.3$\pm$17.7&128.1$\pm$19.8&\textbf{137.5$\pm$20.1}\\
\midrule
Cooperative Navigation&OMAR&MA-SfBC&MA-FQL&OM\textsuperscript{2}P\\
\midrule
Medium-Replay&{260.7$\pm$37.7}&196.1$\pm$11.1&{375.6$\pm$41.7}&\textbf{380.3$\pm$61.8}\\
Medium&{348.7$\pm$51.7}&276.3$\pm$8.8&\textbf{469.6$\pm$42.1}&441.2$\pm$21.3\\
Medium-Expert&450.3$\pm$39.0&299.8$\pm$16.8&\textbf{525.0$\pm$25.9}&510.0$\pm$14.5\\
Expert&564.6$\pm$8.6&553.0$\pm$41.1&{582.6$\pm$26.0}&\textbf{614.0$\pm$4.8}\\
\midrule
Avg.&406.1$\pm$34.2&331.3$\pm$19.4&\textbf{488.2$\pm$33.9}&\textbf{486.4$\pm$25.6}\\
\midrule
HalfCheetah-v2&OMAR&MA-SfBC&MA-FQL&OM\textsuperscript{2}P\\
\midrule
Medium-Replay&1674.8$\pm$201.5&-128.3$\pm$71.3&\textbf{2553.6$\pm$337.4}&2525.9$\pm$441.3\\
Medium&\textbf{2797.0$\pm$445.7}&1386.8$\pm$248.8&{2172.3$\pm$242.3}&2673.9$\pm$225.1\\
Medium-Expert&{2900.2$\pm$403.2}&1392.3$\pm$190.3&{2655.2$\pm$368.6}&\textbf{2917.0$\pm$301.2}\\
Expert&2963.8$\pm$410.5&2386.6$\pm$440.3&{3869.8$\pm$329.3}&\textbf{4151.6$\pm$275.0}\\
\midrule
Avg.&2583.9$\pm$365.2&1259.3$\pm$237.7&{2812.7$\pm$319.4}&\textbf{3067.1$\pm$310.7}\\
\bottomrule
\end{tabular}
\end{center}
\caption{Comparative performance of OM\textsuperscript{2}P against OMAR, MA-SfBC and MA-FQL under different tasks and datasets. Values within \textbf{1\%} of the best performance in each row are highlighted in bold.}
\label{tab:algorithmsmpe}
\end{table*}

\section{Experiments}
We assess the effectiveness of our proposed method across diverse multi-agent environments and datasets. Our evaluation emphasizes these key aspects:  
(1) \textbf{Performance} — assessing how OM\textsuperscript{2}P compares to existing state-of-the-art offline MARL algorithms in standard benchmarks;  
(2) \textbf{Efficiency} — examining whether our algorithm can maintain competitive performance while reducing memory consumption during training, and whether it can leverage efficient one-step sampling to generate high-quality policies during both training and inference.

\subsection{Experimental Setup}

\paragraph{Environments.}  
We perform comprehensive evaluations of our method on two commonly used benchmarks for multi-agent reinforcement learning: the Multi-Agent Particle Environment (MPE) \cite{lowe2017multi} and the Multi-Agent MuJoCo (MAMuJoCo) suite \cite{peng2021facmac}.  
For our experiments, we choose the Predator-Prey, World, and Cooperative Navigation scenarios from MPE, and a two-agent HalfCheetah setup from MAMuJoCo.  
Details of the environment descriptions about the mechanisms are included in Appendix~\ref{appendix:envdescription}.

\paragraph{Datasets.}  
Following the dataset generation protocol proposed in \cite{fu2020d4rl}, we construct four types of datasets to reflect a range of policy qualities. These include medium-replay, medium, medium-expert, and expert datasets. Each type captures different levels of behavior performance and data coverage. Full descriptions of the dataset construction process and characteristics are listed below:  
\begin{itemize}
    \item \textbf{Medium-replay}: includes all transitions stored in the replay buffer during training until the policy reaches medium-level performance.
    \item \textbf{Medium}: consists of $1$ million transitions generated by unrolling a policy whose performance corresponds to the medium level.
    \item \textbf{Expert}: consists of $1$ million transitions collected by executing a well-trained expert policy.
    \item \textbf{Medium-expert}: formed by mixing medium and expert data in fixed proportions ($90\%$ medium and $10\%$ expert for MPE; $99.9\%$ medium and $0.1\%$ expert for MAMuJoCo).
\end{itemize}

\paragraph{Network Structures.}\label{appendix:hyperparameters}

In OM\textsuperscript{2}P, we employ a multi-layer perceptron (MLP) to parameterize the mean-velocity function, which serves as the core of our policy network. The MLP takes the observation, noise and timestep concatenation as inputs and outputs the estimated mean-velocity vector used for one-step action generation. This design offers a lightweight and efficient architecture, enabling faster inference and lower memory consumption. All critic networks, including those in baseline methods such as OMAR and MA-FQL~\cite{pan2022plan,park2025flowqlearning}, similarly adopt MLPs that map concatenated state-action pairs to scalar Q-values. Unless otherwise noted, all networks use GeLU activation and layer normalization after activation, four hidden layers with 512 neurons, and a consistent feature dimension across agents.

\paragraph{Hyperparameters.}

Across all tasks in MPE and MAMuJoCo, we use a unified learning rate of $3\times 10^{-4}$ for both the Q-function and the mean-flow policy network. Due to variations in dataset quality and underlying data distributions, we adopt different values for the Q-weight and the timestep sampling distribution. The specific configurations used for each task are provided in Table~\ref{tab:threshold}. Other fixed hyperparameters are shown in Table~\ref{tab:hyperparameters}.

\begin{table}[ht]
\caption{The $\eta$ value and the $\xi$ value in OM\textsuperscript{2}P.}
\label{tab:threshold}
\begin{center}
\begin{tabular}{cccc}
\toprule
Predator Prey&$\eta$&$\xi$\\
\midrule
Medium-Replay&$100.0$&$[0,0,0,0]$\\
Medium&$10.0$&$[0,0,0.2,0]$\\
Medium-Expert&$100.0$&$[2,0,0,0]$\\
Expert&$1.0$&$[0,0,0,0]$\\
\midrule
World&$\eta$&$\xi$\\
\midrule
Medium-Replay&$100.0$&$[0,0,0.2,0]$\\
Medium&$1000.0$&$[0,0,0.2,0]$\\
Medium-Expert&$10000.0$&$[0,0,0,0]$\\
Expert&$1.0$&$[0,0,0,0]$\\
\midrule
Cooperative Navigation&$\eta$&$\xi$\\
\midrule
Medium-Replay&$1.0$&$[0,0,0,0]$\\
Medium&$1.0$&$[0,0,0,0]$\\
Medium-Expert&$0.1$&$[0,0,0.05,0]$\\    
Expert&$1.0$&$[0,0,0.1,0]$\\
\midrule
HalfCheetah-v2&$\eta$&$\xi$\\
\midrule
Medium-Replay&$1.0$&$[0,0,0,0]$\\
Medium&$0.125$&$[0,0,0,0]$\\
Medium-Expert&$0.02$&$[0,0,0,0]$\\
Expert&$0.008$&$[0,0,0,0]$\\
\bottomrule
\end{tabular}
\end{center}
\end{table}

\paragraph{Baselines.}  
We benchmark our proposed OM\textsuperscript{2}P algorithm against several representative offline multi-agent reinforcement learning methods, including OMAR \cite{pan2022plan}, a multi-agent adaptation of the diffusion-based SfBC policy \cite{chenoffline} (denoted MA-SfBC), and a multi-agent adaptation of flow-based FQL \cite{park2025flowqlearning} (denoted MA-FQL).  
These methods are implemented based on decentralized training with independent actor-critic structures. Besides, we compare our algorithm with MADIFF\cite{zhu2024madiff}, CFCQL\cite{shao2023counterfactual} and DoF\cite{lidof2025} as a centralized training decentralized execution algorithm in MPE tasks (Due to the dataset inconsistency, the results are shown in Appendix~\ref{appendix:expbaselines}).

\paragraph{Devices and Experimental Details.} We run all algorithms on a single NVIDIA RTX A6000 GPU. To ensure statistical reliability, we conduct each algorithm using five different random seeds and report the mean and standard deviation of the evaluation returns. Each model is trained for 1 million timesteps, with evaluations performed every 100 timesteps. During each evaluation, the policy is tested over 10 episodes, and the final performance is reported as the mean and standard deviation across the five seeds at the same timestep. For further explanations about the experimental results, see Appendix~\ref{appendix:expresults} for details.

\begin{table}[ht]
\caption{Hyperparameters in OM\textsuperscript{2}P.}
    \centering
    \begin{tabular}{cc}
    \toprule
    Hyperparameter & Value\\
    \midrule
    Batch Size& $256$\\
    \midrule
    Discount Factor $\gamma$ & $0.99$ \\
    \midrule
    Target Network Update Rate $\rho$ & $0.005$ \\
    \midrule
    Temporal Difference $\Delta r$ & $10^{-12}$ \\
    \bottomrule
    \end{tabular}
    
    \label{tab:hyperparameters}
\end{table}

\vspace{-0.2in}

\subsection{Empirical Results}
\paragraph{Standard Tasks and Datasets.} OM\textsuperscript{2}P consistently achieves superior performance across both MPE and MA-MuJoCo benchmarks, as shown in Table~\ref{tab:algorithmsmpe}. It outperforms OMAR and MA-SfBC by substantial margins and matches or surpasses MA-FQL in most scenarios. Thanks to its one-step policy design, OM\textsuperscript{2}P demonstrates stable learning across datasets of varying quality, while significantly improving training and inference efficiency. On the more challenging multi-agent HalfCheetah-v2 domain, OM\textsuperscript{2}P maintains strong scalability and robustness, achieving nearly the highest average return under all dataset regimes and excelling particularly on expert data. These results collectively highlight OM\textsuperscript{2}P’s effectiveness as a scalable, efficient, and high-performing solution for offline multi-agent reinforcement learning.

\paragraph{Training Efficiency.} To justify the effectiveness of OM\textsuperscript{2}P algorithm, we evaluate the training efficiency of OM\textsuperscript{2}P in terms of GPU memory usage and training time over $10,000$ steps, evaluating $10$ episodes per $100$ steps under the MPE World task using the expert dataset, which is shown in Table~\ref{tab:efficiency}. 
Notably, OM\textsuperscript{2}P maximizes efficiency via derivative-free, native one-step generation, bypassing the high memory costs and distillation overheads typical of baselines (MA-FQL).
Specifically, OM\textsuperscript{2}P reduces GPU memory usage by over $37\%$ (from $1036$MB to $650$MB) and cuts training time by more than $90\%$ (from $5674$s to $564$s for $10.1\times$ improvement) relative to the diffusion-based MA-SfBC. It also achieves over $28\%$ lower memory usage ($906$MB to $650$MB) and approximately $50\%$ faster evaluation ($169$s to $85$s) than the flow-based MA-FQL.
Furthermore, when replacing our derivative-free approximation with full gradient computation, memory usage jumps to over $2.4$GB—demonstrating the importance of our efficient training design to achieve $3.8\times$ GPU memory reduction. 
The results demonstrate that OM\textsuperscript{2}P achieves consistently low memory overhead and fast runtime, highlighting its effectiveness in resource-constrained environments.

\begin{table}[ht]
\caption{Training and inference efficiency comparison via the GPU memory usage, training time (training $10,000$ steps) and evaluating time (summation of evaluating $10$ episodes per $100$ steps during the process of training $10000$ steps) across different methods and ablations with exact gradient value under the MPE World task using the expert dataset.}
\centering
\begin{tabular}{lccc}
\toprule
\textbf{Algorithm} & \textbf{GPU Use} & \textbf{Training time} & \textbf{Evaluation time} \\
\midrule
MA-SfBC & $1036$MB & $5674$s &$975$s\\
MA-FQL & $906$MB & $585$s &$169$s\\
OM\textsuperscript{2}P w/ grad. & $2442$MB & $633$s &\bm{$81$}s \\
\textbf{OM\textsuperscript{2}P (ours)} & \bm{$650$}MB & \bm{$564$}s &\bm{$85$}s \\
\bottomrule
\end{tabular}
\label{tab:efficiency}
\end{table}

\paragraph{Scalability.} To evaluate the scalability of our framework, we conduct our experiments on the Cooperative Navigation task for large number of agents (The baseline task has three agents). The datasets are collected using the policies trained by HATD3\cite{JMLR:v25:23-0488} algorithm for tasks with more agents. The dataset has 40000 trajectories and each trajectory has $25$ timesteps. Table \ref{tab:scale} below shows that OM\textsuperscript{2}P algorithm outperforms other baseline algorithms, which means that our algorithm holds strong scalability among more agents.

\begin{table}[ht]
\caption{Performance comparison with more agents in Cooperative Navigation tasks with $4$ and $5$ agents.}
\centering
\begin{tabular}{lcc}
\toprule
\textbf{Algorithm} & \textbf{4 Agents} & \textbf{5 Agents}\\
\midrule
OMAR & 447.1$\pm$68.1 & 379.8$\pm$66.9 \\ 
MA-SfBC & 450.6$\pm$43.1 & 367.0$\pm$47.8 \\
MA-FQL & 586.1$\pm$43.4 & 615.7$\pm$46.9 \\
\textbf{OM\textsuperscript{2}P (Ours)} & \textbf{587.2$\pm$26.9} & \textbf{621.8$\pm$70.1} \\
\bottomrule
\end{tabular}
\label{tab:scale}
\end{table}

Overall, the results demonstrate the advantages of OM\textsuperscript{2}P across benchmarks. It consistently matches or surpasses diffusion- and flow-based baselines while maintaining stable performance across datasets of varying quality. Its one-step mean-flow design yields substantial efficiency gains, reducing GPU memory and cutting training time up to tenfold compared to prior generative approaches. OM\textsuperscript{2}P also scales effectively to larger agent populations, outperforming strong baselines in cooperative tasks with four and five agents. These findings highlight OM\textsuperscript{2}P’s combination of high performance, efficiency, and scalability.

\subsection{Ablation Study}
We perform ablation studies to investigate these key aspects: the sensitivity of critical hyperparameters, the isolated contributions of core components within OM\textsuperscript{2}P, the robustness across dataset sizes, and the scalability to large numbers of agents.

\begin{figure}[ht]
\centering
\includegraphics[width=\linewidth]{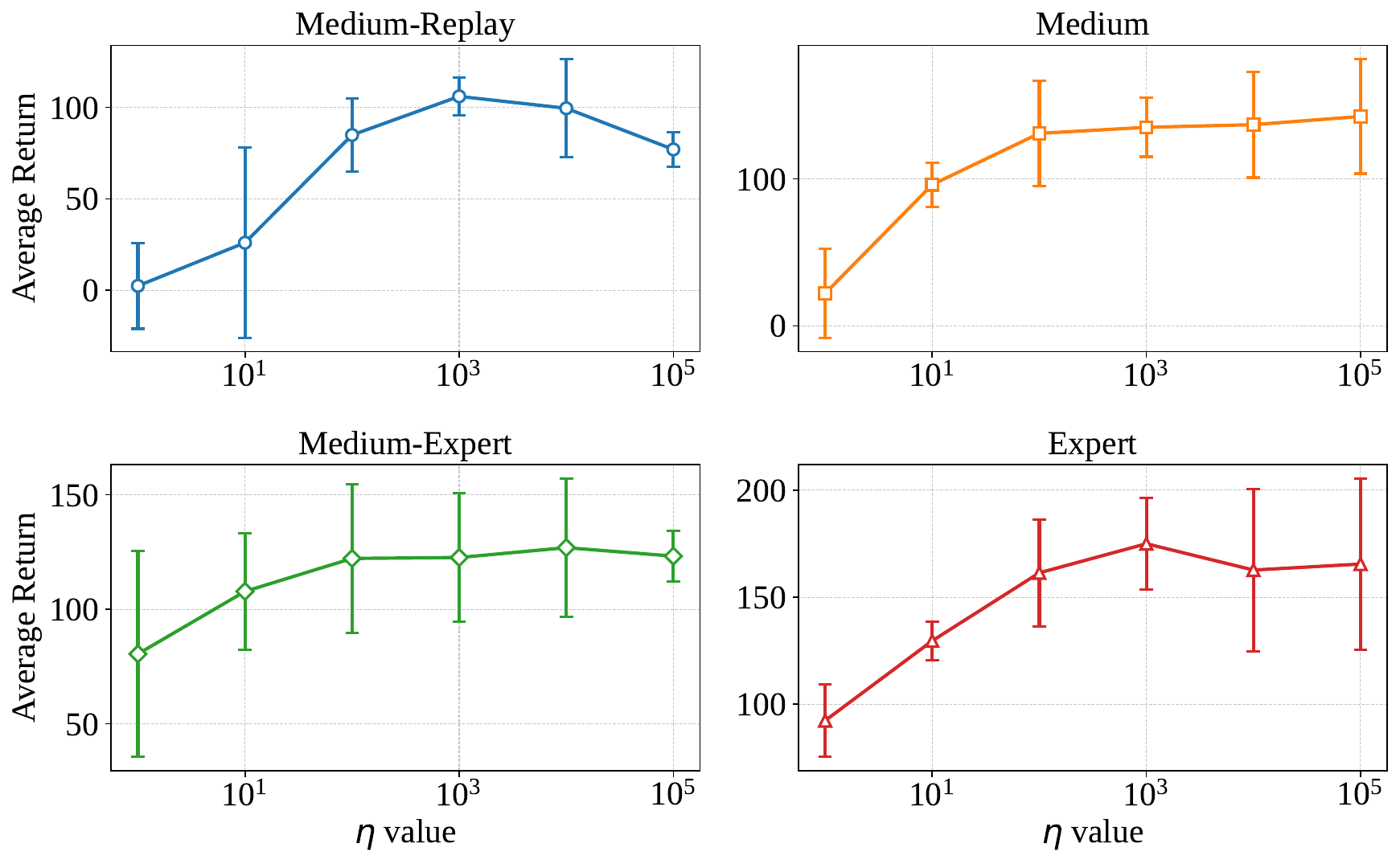}
\vspace{-0.2in}
\caption{Performance impact of the coefficient $\eta$ in MPE World.}
\label{fig:ablationstudy_qweight}
\end{figure}

\vspace{-0.2in}

\paragraph{Hyperparameters.} We examine the role of the mixing coefficient $\eta$, which balances the behavior cloning (BC) loss and the Q-learning objective. Figure~\ref{fig:ablationstudy_qweight} presents the performance of our method across four datasets in the MPE World task under varying $\eta$ values. The results indicate that $\eta$ is a crucial hyperparameter that must be tuned in accordance with the quality of the offline dataset to achieve optimal results. Notably, small $\eta$ values lead to suboptimal learning due to insufficient Q-value guidance, while moderate to large values (e.g., $10^3$) 
consistently deliver stronger performance, particularly on high-quality datasets. This underscores the importance of properly calibrating $\eta$ to strike a balance between policy imitation and reward-driven optimization.

\begin{figure}[ht]
\centering
\includegraphics[width=0.8\linewidth]{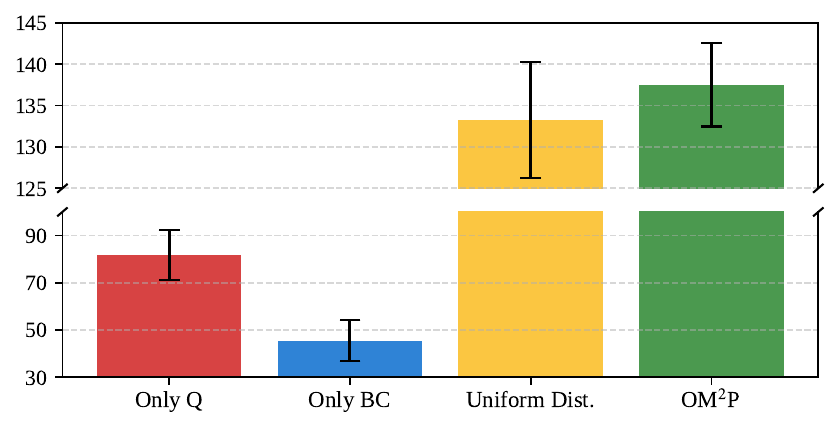}
\vspace{-0.2in}
\caption{Effect of removing key components from OM\textsuperscript{2}P on the MPE World task. Each ablation variant shows reduced average returns compared to the full model, confirming the importance of each module.}
\label{fig:ablationstudy_component}
\end{figure}

\paragraph{Ablation on different components.} To assess the contribution of each key component in OM\textsuperscript{2}P, we conduct modular ablations by systematically removing or modifying core elements of the framework. Specifically, we evaluate three simplified variants: using only the Q-learning objective (removing the BC term), using only behavior cloning (removing the value-guided objective), and replacing the general timestep distribution with a uniform distribution. As shown in Figure~\ref{fig:ablationstudy_component}, each ablation leads to a noticeable drop in performance compared to the full OM\textsuperscript{2}P. 
These results collectively demonstrate that the synergy among all components—Q supervision, BC guidance, and timestep reweighting—is critical to realizing the full potential of OM\textsuperscript{2}P.

\paragraph{Robustness across dataset scales.}
To further evaluate the robustness of our method, we conduct ablation studies on different sizes of offline datasets under the \textit{World} task expert dataset. As shown in Table~\ref{tab:dataset_ablation}, OM\textsuperscript{2}P consistently maintains high performance even when the dataset size is significantly reduced, demonstrating strong robustness and data efficiency in offline MARL.

\begin{table}[h]
\centering
\caption{Performance of OM$^2$P under different dataset sizes on the \textit{World} task expert dataset.}
\label{tab:dataset_ablation}
\begin{tabular}{cccc}
\toprule
Number of Trajectories & 4e4 & 3.5e4 & 3e4 \\
\midrule
Performance of OM\textsuperscript{2}P & 175.0$\pm$21.5 & 160.2$\pm$5.2 & 154.3$\pm$10.4 \\
\bottomrule
\end{tabular}
\end{table}

The ablation results highlight the effectiveness and robustness of OM\textsuperscript{2}P. Proper tuning of the mixing coefficient $\eta$ is essential to balance policy imitation and reward guidance, with moderate to large values yielding consistently strong performance. Modular ablations confirm that each core component—Q-function supervision, behavior cloning, and generalized timestep weighting—contributes substantially to overall performance, and removing any element degrades returns. Furthermore, OM\textsuperscript{2}P maintains high performance even under reduced dataset sizes, demonstrating strong data efficiency and robustness. Collectively, these findings validate the design choices of OM\textsuperscript{2}P and underscore its ability to deliver stable, efficient, and scalable offline multi-agent reinforcement learning.

\section{Conclusion}
In this paper, we propose OM\textsuperscript{2}P, a novel offline multi-agent RL algorithm that integrates mean-flow model into one-step policy generation. OM\textsuperscript{2}P addresses core limitations of existing generative methods by aligning training with reward-based objectives, introducing adaptive timestep sampling, and employing a memory-efficient derivative-free velocity estimation. Extensive experiments on MPE and MAMuJoCo demonstrate that OM\textsuperscript{2}P achieves state-of-the-art performance with significantly improved efficiency. OM\textsuperscript{2}P efficiently integrates mean-flow model into offline MARL, enabling scalable, efficient, and high-quality policy learning in complex multi-agent settings.

\section{Acknowledgement}
This work is supported by the National Natural Science Foundation of China Grant 52494974.

\bibliographystyle{plain}
\bibliography{plain} 
\section{Appendix}

\begin{figure*}[ht]
\begin{center}
\centering
\subfloat[Cooperative Navigation]{
\includegraphics[width=0.25\textwidth]{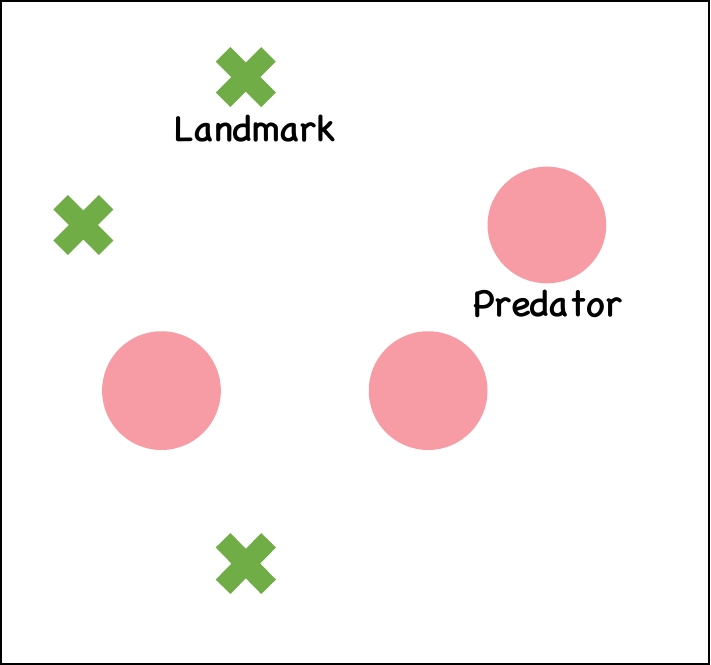}
\label{Fig3-1}
}
\quad
\subfloat[Predator Prey]{
\includegraphics[width=0.25\textwidth]{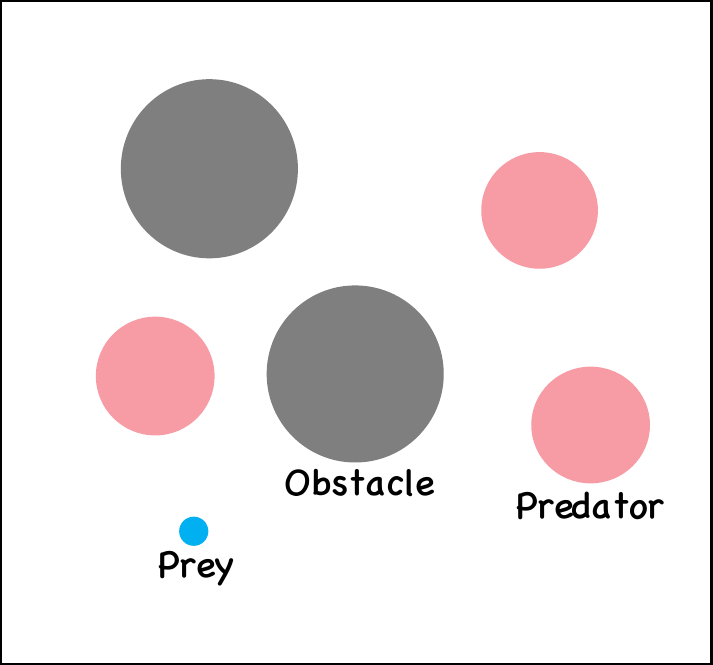}
\label{Fig3-2}
}
\quad
\subfloat[World]{
\includegraphics[width=0.25\textwidth]{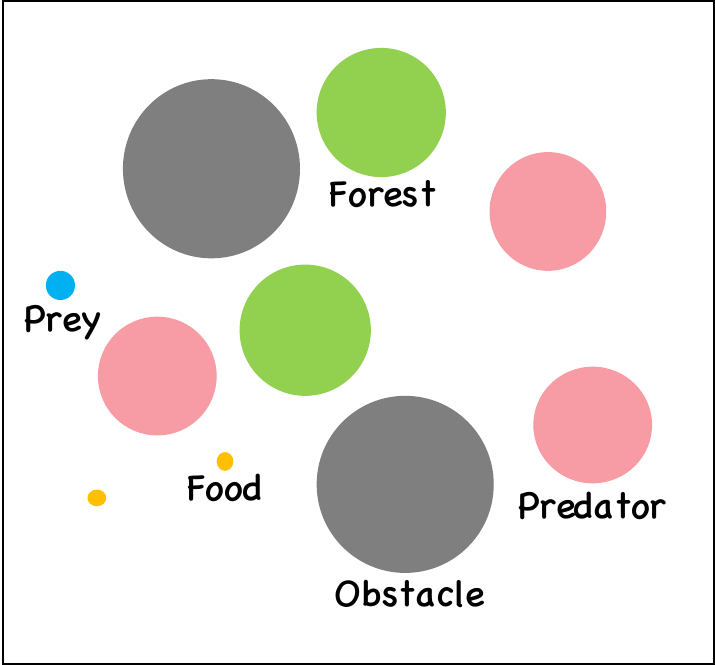}
\label{Fig3-3}
}
\quad
\subfloat[2-Agent HalfCheetah]{
\includegraphics[width=0.35\textwidth]{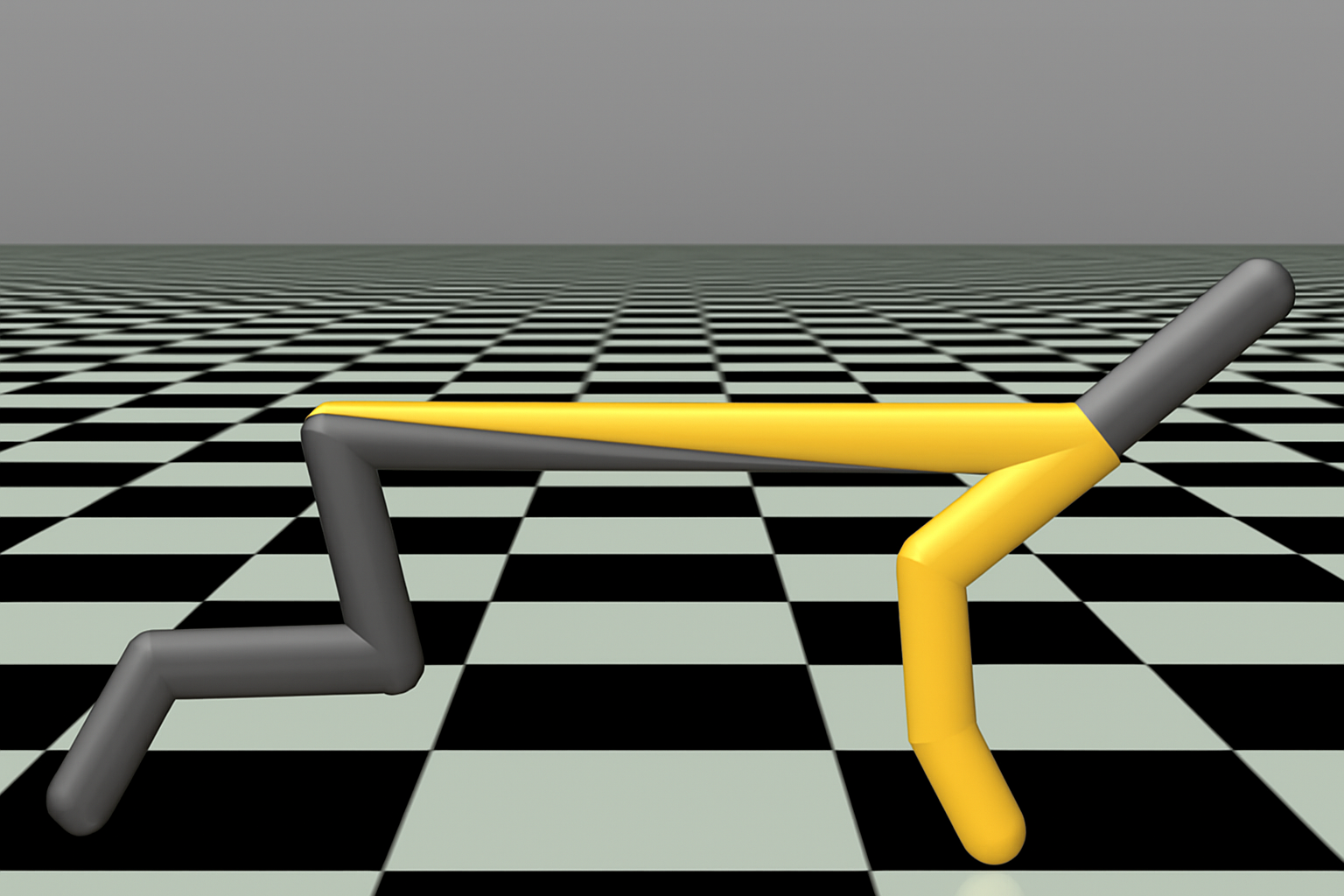}
\label{Fig3-4}
}

\caption{Multi-agent particle environments (MPE) and Multi-agent HalfCheetah task in MuJoCo Environment (MAMuJoCo).}
\label{fig:environments}
\end{center}
\end{figure*}

\subsection{Experimental Details: Environments Description}\label{appendix:envdescription}

We build our implementation of OM\textsuperscript{2}P and all baseline methods upon two widely used open-source simulation platforms: the Multi-Agent Particle Environment (MPE) \cite{lowe2017multi}\footnote{https://github.com/openai/multiagent-particle-envs} and the Multi-Agent MuJoCo suite (MAMuJoCo) \cite{peng2021facmac}\footnote{https://github.com/schroederdewitt/multiagent\_mujoco}. MPE consists of physical environments where agents must coordinate to accomplish cooperative objectives. In contrast, MAMuJoCo extends the classic MuJoCo locomotion tasks into a multi-agent form, where control of a robot is distributed across several agents. The task setups from both environments are illustrated in Figure~\ref{fig:environments}.

In the Cooperative Navigation task (Figure~\ref{Fig3-1}), multiple agents (pink circles) are required to navigate toward target landmarks (green crosses) while avoiding collisions. The Predator-Prey scenario (Figure~\ref{Fig3-2}) involves pink agents (predators) working together to capture a blue agent (prey), all while avoiding static obstacles (grey cricles). Notably, predators are slower than the prey, necessitating coordinated movement for successful capture.

The World task (Figure~\ref{Fig3-3}) includes three slower cooperative agents (pink) pursuing a single faster adversary (blue), who seeks to reach food items (orange dots). Agents must avoid colliding with obstacles (grey) and deal with the additional challenge that the adversary can obscure itself in forests (light green regions), making its location partially unobservable.

Lastly, the 2-Agent HalfCheetah task (Figure~\ref{Fig3-4}) requires two agents to jointly control distinct limbs (grey and yellow segments, regardless of the head) of a cheetah body in order to achieve fast and stable forward locomotion.

\begin{table*}[h]
\centering
\caption{Comparative performance of OM\textsuperscript{2}P against OMAR, MA-SfBC, CFCQL, MADIFF, DoF and MA-FQL under different tasks and datasets of Multi-Agent Particles Environment (MPE). Values within \textbf{1\%} of the best performance in each row are highlighted in bold.}
\label{tab:algorithmsmpeotherbaselines}
\resizebox{\textwidth}{!}{
\begin{tabular}{cccccccc}
\toprule
Predator–Prey&OMAR&MA-SfBC&CFCQL&MADIFF&DoF&MA-FQL&OM\textsuperscript{2}P\\
\midrule
Medium-Replay&{86.8$\pm$43.7}&26.1$\pm$10.0&130.8$\pm$11.4&114.1$\pm$17.5&94.0$\pm$19.2&\textbf{203.8$\pm$43.3}&\textbf{203.1$\pm$59.3}\\
Medium&116.9$\pm$45.2&127.0$\pm$50.9&125.8$\pm$41.4&142.3$\pm$19.7&155.1$\pm$18.2&{255.1$\pm$65.0}&\textbf{257.3$\pm$52.0}\\
Expert&202.8$\pm$27.1&256.0$\pm$26.9&220.1$\pm$24.9&225.2$\pm$27.7&223.7$\pm$12.0&{289.5$\pm$28.6}&\textbf{330.3$\pm$40.9}\\
\midrule
World&OMAR&MA-SfBC&CFCQL&MADIFF&DoF&MA-FQL&OM\textsuperscript{2}P\\
\midrule
Medium-Replay&21.1$\pm$15.6&9.1$\pm$5.9&56.5$\pm$20.0&42.5$\pm$9.2&43.3$\pm$9.9&\textbf{107.0$\pm$30.0}&\textbf{106.0$\pm$10.2}\\
Medium&45.6$\pm$16.0&54.2$\pm$22.7&74.1$\pm$27.4&99.8$\pm$3.9&67.8$\pm$9.1&136.1$\pm$18.5&\textbf{142.0$\pm$18.6}\\
Expert&84.8$\pm$21.0&97.3$\pm$19.1&96.5$\pm$22.8&99.0$\pm$12.4&112.6$\pm$17.3&139.4$\pm$13.9&\textbf{175.0$\pm$21.5}\\
\midrule
Cooperative Navigation&OMAR&MA-SfBC&CFCQL&MADIFF&DoF&MA-FQL&OM\textsuperscript{2}P\\
\midrule
Medium-Replay&{260.7$\pm$37.7}&196.1$\pm$11.1&346.2$\pm$34.3&268.0$\pm$8.9&331.5$\pm$12.9&{375.6$\pm$41.7}&\textbf{380.3$\pm$61.8}\\
Medium&{348.7$\pm$51.7}&276.3$\pm$8.8&391.9$\pm$36.4&391.5$\pm$27.4&375.8$\pm$30.3&\textbf{469.6$\pm$42.1}&441.2$\pm$21.3\\
Expert&564.6$\pm$8.6&553.0$\pm$41.1&559.6$\pm$14.3&498.9$\pm$18.9&610.7$\pm$11.1&{582.6$\pm$26.0}&\textbf{614.0$\pm$4.8}\\
\bottomrule
\end{tabular}
}
\end{table*}

\subsection{Details about the Experimental Results}\label{appendix:expresults}
The results presented in Figure~\ref{fig:generalized_bc} are obtained on the MPE World task using the expert dataset. In the left subfigure, the model is trained solely using the mean-flow loss, whereas the right subfigure corresponds to training with the full OM\textsuperscript{2}P algorithm and the calculation of the target mean velocity differs from each other to clarify the efficiency of deriative-free estimation and the appropriate range of the temporal difference value $\Delta r$ to guarantee the numerical stability.

Figure~\ref{fig:ablationstudy_component} reports the average performance and standard deviation across four datasets of varying quality in the MPE World task, where the final scores are computed by averaging results obtained on each dataset.

We adopt the expert and random mean episode returns from each MPE task, with the expert/random returns being $\{516.8,159.8\}$ for Cooperative Navigation, $\{185.6,-4.1\}$ for Predator-Prey, and $\{79.5,-6.8\}$ for World.

\subsection{Comparisons with baseline algorithms}\label{appendix:expbaselines}

Our OM\textsuperscript{2}P algorithm is designed as a decentralized method with a strong emphasis on training efficiency. Although a fully fair comparison across methods is inherently challenging, we benchmark OM\textsuperscript{2}P against representative baselines, including CFCQL \cite{shao2023counterfactual}, MADIFF \cite{zhu2024madiff}, and DoF \cite{lidof2025}. The results, summarized in Table~\ref{tab:algorithmsmpeotherbaselines}, demonstrate that OM\textsuperscript{2}P consistently outperforms these methods. To further ensure comparability, all evaluations are conducted on three dataset variants excluding the medium-expert setting, thereby standardizing the difficulty level across algorithms.

We finally discuss and clarify that OM\textsuperscript{2}P is a specialized decentralized framework designed for continuous control, distinct from Transformer-based baselines such as MADT \cite{meng2023offline} which rely on next-token prediction optimized for discrete action spaces. Regarding efficiency, OM\textsuperscript{2}P achieves a relative advantage over methods like FQL by natively enabling one-step mean-flow generation, thereby bypassing iterative overhead.  Figure~\ref{fig:om2p} illustrates this scalable agent-level mechanism rather than a direct adaptation of single-agent models. Finally, establishing formal convergence guarantees and integrating explicit conservative mechanisms to further mitigate overestimation risks represent valuable directions for future theoretical expansion.

\end{document}